\documentclass[11pt]{article}

\usepackage[final]{acl}

\usepackage{times,graphicx,tabularx}
\usepackage{latexsym,amsmath,bbm}
\usepackage{longtable}
\usepackage{booktabs}
\usepackage{xltabular}
\usepackage{url}
\usepackage[T1]{fontenc}

\usepackage[utf8]{inputenc}

\usepackage{microtype}

\usepackage{inconsolata}

\usepackage{graphicx}

\title{SN-WER: Script-Normalized WER for Multi-Script Indic ASR Evaluation}

\author{Priyaranjan Pattnayak \\
  Oracle America Inc. \\
  \texttt{priyaranjanpattnayak@gmail.com}}

\begin{document}
%
\maketitle
\begin{abstract}

Word Error Rate (WER) is the dominant ASR metric but can overestimate errors when references and hypotheses encode the same words in different scripts, a common issue in multilingual settings where models emit romanized text. We propose \textbf{Script-Normalized WER (SN-WER)}, a \textbf{training-free, evaluation-only} scoring method that transliterates reference and hypothesis into a language-specific canonical script before computing WER. Evaluated on \textbf{5 Indic languages}, \textbf{2 datasets}, and \textbf{3 ASR models}, SN-WER quantifies script mismatch effects: on curated FLEURS it narrows inflated model gaps by up to \textbf{12\%}, while on noisy Common Voice the reductions are smaller or inconsistent, exposing genuine recognition weaknesses rather than only script mismatch. Controlled stress tests show \textbf{67\% attenuation} of artificial romanization-induced WER inflation, while lexical-substitution controls show near-identical sensitivity to semantic errors (\(\Delta\)SN/\(\Delta\)WER \(\approx \textbf{1.09}\)). SN-WER is robust to transliterator choice ($\Delta<0.002$), normalization changes ($\Delta<0.05$), and has low token-collision rates (\(<\textbf{0.1\%}\)) in the evaluated Indic setting. We argue that SN-WER should be reported alongside WER/CER as a companion score for script-insensitive ASR evaluation, especially when transcripts feed downstream search, indexing, or multilingual LLM pipelines.

\end{abstract}
%
\section{Introduction}
\label{sec:intro}
\noindent \textbf{Motivation:}
Multilingual ASR increasingly serves low-resource communities, yet evaluation still hinges on \emph{Word Error Rate} (WER)~\cite{morris2004wer}. WER assumes orthographic consistency between reference and hypothesis; however, multilingual settings span languages written in diverse scripts, and some admit multiple or informal \emph{romanization conventions}. Prior work shows that normalization choices and orthographic variation can substantially alter reported error rates in multilingual ASR evaluation~\cite{manohar2024pitfalls, hunt1990effects, thennal2025cer}. 
When script differences are treated as lexical substitutions, WER can inflate apparent error, complicating cross-language comparison, with particular impact on Indic languages, represented here across five distinct scripts.


\noindent \textbf{Evidence of script bias.}

On FLEURS~\citep{conneau2023fleurs}, several Indic-language outputs contain romanized tokens despite native-script references. For example, Whisper~\citep{radford2023whisper} on Odia gives WER \(=1.13\), while script-normalized scoring reduces it to \(1.02\), indicating a measurable script-mismatch component. Aggregated across curated FLEURS, SN-WER narrows inflated model gaps by up to \textbf{12\%}. On noisy Common Voice~\citep{ardila2020common}, reductions are smaller or less consistent, suggesting that noisy conditions contain genuine recognition errors rather than only script mismatch. Thus, SN-WER is useful as a companion score: it separates surface-script penalties from lexical recognition errors in settings where script choice is not the target of evaluation.

\noindent \textbf{Related Work:}

Transliteration-optimized WER (\emph{toWER}) was introduced for \emph{code-switched} Indic speech, mapping text to a shared writing system and showing that evaluation outcomes can shift under transliteration~\cite{emond2018tower}. Our focus differs in three ways: (i) \emph{monolingual cross-script evaluation}, especially for Indic languages where romanized output is common across multiple scripts (Devanagari, Bengali, Tamil, Gujarati, Odia); (ii) \emph{evaluation-only scoring}, requiring no retraining or decoding changes; and (iii) systematic robustness analysis across transliterators, normalization choices, adversarial sanity checks, bootstrap CIs, and curated vs.\ noisy benchmarks.

Other metrics, such as WERd~\cite{ali2019mgb} for dialectal Arabic and lenient CER~\cite{karita2023lenient} for Japanese, address spelling-variant classes or character-level inconsistencies rather than cross-script canonicalization. Recent work also shows CER as a more consistent alternative to WER in multilingual ASR evaluation~\cite{thennal2025cer}. These efforts highlight the importance of normalization~\cite{manohar2024pitfalls}, but do not directly evaluate monolingual multi-script Indic settings where native-script references are compared against romanized hypotheses.

We do not claim transliteration-before-scoring as a new idea. Instead, SN-WER provides a focused evaluation-only formulation for script-normalized WER in multi-script Indic ASR. Our contribution is empirical and methodological: we quantify script-mismatch effects across five Indic scripts, compare curated and noisy datasets, test robustness across transliteration choices, measure collision risk, and use controlled perturbations to distinguish script mismatch from lexical error. SN-WER should therefore be reported alongside WER/CER as a companion score, not as a replacement.

\paragraph{Scope.}
SN-WER is an evaluation-only companion to WER/CER, not a replacement for them. It is appropriate when script choice is orthogonal to the downstream task, such as keyword search, indexing, retrieval, intent classification, or downstream multilingual LLM processing. For user-facing transcripts, captions, subtitles, or educational applications, correct script choice remains part of output quality; in those settings, WER/CER and SN-WER should be reported together.



\noindent \textbf{Our Contributions:}
\begin{itemize}
    \item \textbf{SN-WER:} an evaluation-only, script-normalized extension of WER for multi-script ASR, requiring no retraining, decoding changes, or additional labeled data.
    
    \item \textbf{Systematic evaluation:} a focused study of script mismatch across \textbf{5 Indic languages}, \textbf{3 ASR models}, and two benchmarks, with additional validation on Arabic and Urdu.
    
    \item \textbf{Script-mismatch quantification:} SN-WER narrows inflated model gaps by up to \textbf{12\%} on curated FLEURS, preserves genuine weaknesses on noisy Common Voice, and yields moderate Arabic/Urdu reductions of \textbf{5--9\%}.
    
    \item \textbf{Evaluation rigor:} robustness across transliterators \((\Delta<0.002)\) and normalization choices \((\Delta<0.05)\), plus romanization-rate correlation, bootstrap CIs, controlled stress tests, and adversarial checks showing sensitivity to semantic errors.
    
    \item \textbf{Impact:} SN-WER complements WER/CER by quantifying measured error attributable to script mismatch, supporting model comparison for script-insensitive uses such as search, indexing, retrieval, and multilingual LLM pipelines.
\end{itemize}

\section{Methodology and Evaluation Framework}

\subsection{Principle}
WER can overestimate lexical error when reference and hypothesis encode the same words in different scripts~\cite{blodgett2020language}. This is common in multi-script ASR evaluation, where benchmarks may use native-script references while models sometimes emit romanized hypotheses. We introduce \textbf{Script-Normalized WER (SN-WER)}, a score that estimates how much of the measured WER is attributable to script mismatch.

\subsection{Properties}
Under a deterministic, boundary-preserving transliteration map, SN-WER has three useful diagnostic properties:

\textbf{Identity.} If reference and hypothesis share the same script, SN-WER $\approx$ WER.  

\textbf{Conservativeness.} If transliteration preserves token boundaries and does not introduce collisions, then script-only mismatches can be removed without increasing edit distance. Under these assumptions, SN-WER is expected to be less than or equal to WER:
\[
\mathrm{SN\text{-}WER}(R,H) \leq \mathrm{WER}(R,H).
\]
In practice, this inequality is not treated as an unconditional guarantee: language-specific mappings may merge distinct forms, split tokens, or introduce transliteration collisions. We therefore report transliterator disagreement and collision rates empirically.
  

\paragraph{Lexical sensitivity.}
SN-WER should not reduce errors caused by incorrect words, deletions, insertions, or word-order changes. We test this empirically with lexical-substitution controls and adversarial sanity checks. These checks verify whether script normalization preserves sensitivity to genuine recognition errors rather than simply lowering scores.

These properties motivate SN-WER as a diagnostic companion to WER, with reliability depending on the transliteration map and preprocessing choices.

\subsection{Definition and Implementation}
Let $R = (r_1,\ldots,r_n)$ be the reference tokens and $H = (h_1,\ldots,h_m)$ the hypothesis.
Standard WER is:
\begin{equation}
\text{WER}(R,H) = \frac{S+D+I}{n},
\end{equation}
where $S,D,I$ are edit operations in Levenshtein alignment.

\begin{equation}
\text{SN-WER}(R,H) = \text{WER}(T(R),T(H)),
\end{equation}
where $T(\cdot)$ maps tokens to a canonical script $C$.

\textbf{Assumptions on $T$:} We assume $T$ is deterministic, boundary-preserving (no token merges/splits), and identity on canonical tokens modulo normalization. Under these conditions, script-only mismatches can be removed without increasing edit distance, so SN-WER$(R,H)$ is expected to be $\leq$ WER$(R,H)$. This is a conditional property rather than an unconditional guarantee: language-specific mappings may merge distinct forms, split tokens, or introduce transliteration collisions. We therefore empirically bound collision effects and transliterator disagreement in Table~\ref{tab:e3-canon-robust}.

We use the benchmark reference script as $C$. For the Indic languages studied here, this corresponds to the native script used in FLEURS and Common Voice. This choice makes SN-WER directly comparable to the benchmark reference while allowing romanized hypothesis tokens to be scored by lexical content rather than surface script. It does not replace orthographic evaluation: when the expected output script is part of the task, standard WER/CER should still be reported.

Implementation uses deterministic transliteration plus standard Unicode, punctuation, and digit normalization. Romanized tokens are detected using Unicode-block heuristics and transliterated into the language-specific reference script. We compare widely used mappings and libraries (ICU, IAST-style, and ITRANS-style where available), showing low disagreement across tools. Experiments with alternative canonical scripts, including Devanagari, shift results by at most $\Delta{<}0.005$.

\subsection{Datasets and Models}
To validate SN-WER, we design a systematic evaluation across curated and noisy benchmarks, multiple language families, and diverse model scales:


\begin{table}[h]
\centering
\small
\setlength{\tabcolsep}{3pt}
\renewcommand{\arraystretch}{1.05}

\begin{tabularx}{\columnwidth}{@{}l X X@{}}
\toprule
\textbf{Dataset} & \textbf{Languages} & \textbf{Models} \\
\midrule
FLEURS 
& hi, bn, ta, or, gu 
& whisper-large-v3, MMS, whisper-small \\

CommonVoice v17 
& hi, bn, ta, or 
& whisper-large-v3, MMS, whisper-small \\

FLEURS Cross-script 
& ar, ur 
& whisper-large-v3, MMS, whisper-small \\
\bottomrule
\end{tabularx}

\caption{Datasets, languages, and models used in evaluation.}
\end{table}

\subsection{Hypotheses}
Our evaluation is structured around four hypotheses:

\noindent\textbf{H1 (Script-mismatch reduction):} SN-WER narrows inflated model gaps on curated datasets by reducing script-mismatch penalties.

\noindent\textbf{H2 (Robustness):} On noisy datasets, SN-WER should not uniformly lower scores; genuine recognition errors should remain visible after script normalization.

\noindent\textbf{H3 (Stability):} SN-WER is stable across transliteration tools and normalization choices.

\noindent\textbf{H4 (Cross-script validation):} SN-WER shows consistent behavior on Indic languages, with additional validation on Arabic and Urdu.


\subsection{Scope and novelty}
SN-WER matches WER’s $O(nm)$ complexity but reframes evaluation by explicitly correcting script bias.

Unlike prior transliteration-based metrics like toWER~\cite{emond2018tower}, which were designed for \emph{code-switched} Indic ASR and sometimes modified training corpora, SN-WER differs in four ways: \textbf{(1) Evaluation-only:} modifies only scoring, requiring no retraining or decoding changes. \textbf{(2) Monolingual multi-script focus:} targets cross-script mismatch in monolingual benchmarks, particularly for Indic languages where romanization is common across five distinct scripts (Devanagari, Bengali, Tamil, Gujarati, Odia). \textbf{(3) Cross-script validation:} evaluated systematically on five Indic languages, with additional validation on Arabic and Urdu, beyond bilingual code-switch settings. \textbf{(4) Diagnostic robustness:} evaluates identity, conditional conservativeness, and lexical sensitivity, and validates robustness across transliterators, normalization choices, adversarial sanity checks, and bootstrap CIs.

By addressing Indic scripts and validating on Arabic and Urdu, SN-WER provides an evaluation-only companion to WER/CER for quantifying script-mismatch effects in multilingual ASR. It is most appropriate when script choice is not part of the downstream task, and should be reported alongside standard WER/CER when orthographic form matters.

\section{Results}

\subsection{E1: Main Effect of SN-WER}
Table~\ref{tab:e1-main} shows average WER and SN-WER across datasets and models. SN-WER reduces inflated gaps by up to 12\% on Gujarati (FLEURS) and 26\% on Odia (Common Voice). MMS outperforms Whisper models, while Whisper-small performs worst, with WER exceeding 1.0 due to heavy insertion errors in low-resource settings.

             

\begin{table}[h]
\centering
\small
\setlength{\tabcolsep}{3pt}
\renewcommand{\arraystretch}{1.05}

\begin{tabular*}{\columnwidth}{@{\extracolsep{\fill}} l l c c c @{}}
\toprule
\textbf{Dataset} & \textbf{Model} & \textbf{WER} & \textbf{SN-WER} & $\Delta$ \\
\midrule
FLEURS & MMS & 0.32 & 0.30 & -5.4 \\
       & Whisper-large & 0.70 & 0.64 & -8.0 \\
       & Whisper-small & 1.27 & 1.21 & -4.7 \\
\midrule
CommonVoice & MMS & 0.46 & 0.36 & -23.0 \\
            & Whisper-large & 0.86 & 0.82 & -4.3 \\
            & Whisper-small & 1.46 & 1.36 & -6.9 \\
\bottomrule
\end{tabular*}

\caption{$\Delta$ is relative change of SN-WER vs. WER, in \%.}
\label{tab:e1-main}

\end{table}


\subsection{E2: Ranking Stability and Leaderboard Impact}
\begin{figure*}[h]
\centering

    \centering
    \includegraphics[width=1\textwidth]{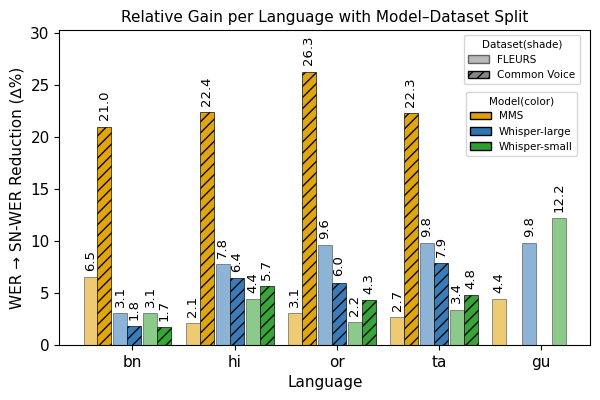}
\caption{Relative WER--SN-WER gap by language, dataset, and model. Relative $\Delta$ \% shrinks for SN-WER.}
\label{fig:leaderboard}

\end{figure*}

Ranks are stable ($\tau=1.0$), but WER-SNWER gap sizes shrinks. On FLEURS, SN-WER narrows MMS–Whisper gaps due to script bias removal; on CV, reductions are minimal due to real errors rather than script mismatch. Fig \ref{fig:leaderboard} shows that SN-WER changes error-gap interpretation without altering rank order while staying noise-sensitive.


\subsection{E3: Canonicalization Robustness}

We evaluate robustness of the canonicalization mapping $T(\cdot)$ across transliteration tools and normalization variants.

\textbf{Tool invariance.} Comparing IAST, ITRANS, and ICU mappings (Table~\ref{tab:e3-translit-indic}) yields negligible differences ($\Delta<0.002$), confirming aggregate invariance.

\textbf{Fine-grained disagreement and collision risk.}
Across native, alt (roman-aware), and ICU pipelines, mean absolute SN-WER disagreement remains $\approx$0.002 (mean P95 $<$ 0.003, Table~\ref{tab:e3-canon-robust}), and fewer than 2-3\% of utterances exceed 0.01 deviation.
Transliteration collision rate (distinct tokens mapping to identical canonical forms) remains below 0.1\% across languages.

\textbf{Normalization robustness.}
Digit and punctuation ablations shift SN-WER by at most $\Delta<0.05$ (Table~\ref{tab:e4-norm-indic}), confirming stability to preprocessing choices.

Together, these results show SN-WER is robust to canonicalization choices and introduces no measurable bias.

\begin{table}[h]
\centering
\begin{tabular}{l c c c}
\toprule
\textbf{Language} & \textbf{IAST} & \textbf{ITRANS} & \textbf{ICU} \\
\midrule
Hindi & 0.421 & 0.420 & 0.421 \\
Bengali & 0.532 & 0.533 & 0.532 \\
Tamil & 0.611 & 0.611 & 0.612 \\
Odia & 0.487 & 0.488 & 0.487 \\
Gujarati & 0.458 & 0.457 & 0.458 \\
\bottomrule
\end{tabular}
\caption{Transliterator invariance on Indic languages ($\Delta{<}0.002$). Values are SN-WER.}
\label{tab:e3-translit-indic}

\end{table}

\begin{table}[h]
\centering
\footnotesize
\setlength{\tabcolsep}{2pt}
\renewcommand{\arraystretch}{1.05}

\begin{tabular*}{\columnwidth}{@{\extracolsep{\fill}} l c c c c @{}}
\toprule
\textbf{Lang} & \textbf{Mean $|\Delta|$} & \textbf{P95} & \textbf{\% $>$ .01} & \textbf{Coll. \%} \\
\midrule
bn & 0.0023 & 0.0032 & 2.79 & 0.032 \\
gu & 0.0018 & 0.0000 & 2.53 & 0.028 \\
hi & 0.0025 & 0.0000 & 2.29 & 0.085 \\
or & 0.0036 & 0.0096 & 3.42 & 0.000 \\
ta & 0.0019 & 0.0000 & 1.77 & 0.006 \\
\midrule
\textbf{Mean} & \textbf{0.0024} & \textbf{0.0026} & \textbf{2.56} & \textbf{0.030} \\
\bottomrule
\end{tabular*}

\caption{Canonicalization robustness across native, alt, and ICU pipelines. Mean absolute SN-WER disagreement $\approx$0.0024; collision rate remains below 0.1\%.}
\label{tab:e3-canon-robust}

\end{table}

\begin{table}[h]
\centering
\begin{tabular}{l c c}
\toprule
\textbf{Language} & \textbf{Base SN-WER} & \textbf{With ablation} \\
\midrule
Hindi & 0.42 & 0.44 \\
Bengali & 0.53 & 0.56 \\
Tamil & 0.61 & 0.65 \\
Odia & 0.49 & 0.52 \\
Gujarati & 0.46 & 0.49 \\
\bottomrule
\end{tabular}
\caption{Normalization ablation on Indic languages}
\label{tab:e4-norm-indic}

\end{table}


\subsection{E4: Cross-Script Extension}


We extend SN-WER to Arabic and Urdu on FLEURS (Table~\ref{tab:cross-script}) as additional validation beyond the main Indic setting. Studies of Urdu ASR models \cite{arif2025wer} already highlighted the need for robust evaluation metrics, as WER alone can misrepresent performance. Both languages show moderate reductions: Arabic improves by 4.9--6.9\%, while Urdu improves by 6.4--9.0\%. These results suggest that script-normalized scoring can be useful beyond Indic when language-specific orthographic conventions are respected.

\begin{table*}[t]
\centering
\small
\setlength{\tabcolsep}{5pt}
\renewcommand{\arraystretch}{1.05}

\begin{tabular*}{0.72\textwidth}{@{\extracolsep{\fill}} l l c c c @{}}
\toprule
\textbf{Lang.} & \textbf{Model} & \textbf{WER} & \textbf{SN-WER} & $\boldsymbol{\Delta}$ \\
\midrule
Arabic   & MMS           & 0.41 & 0.39 & -4.9 \\
         & Whisper-large & 0.16 & 0.14 & -6.9 \\
\midrule
Urdu     & MMS           & 0.31 & 0.29 & -6.4 \\
         & Whisper-large & 0.77 & 0.70 & -9.0 \\
\bottomrule
\end{tabular*}

\caption{Cross-script extension: SN-WER reduces inflated errors in Arabic and Urdu. $\Delta$ is reported in \%.}
\label{tab:cross-script}
\end{table*}


Correction scales with romanization prevalence: Indic shows high inflation, Arabic and Urdu show moderate (5–9\%). This shows SN-WER’s script-specific sensitivity and generalizing ability beyond Indic. While stress tests (E5--E7) focus on Indic languages where romanized outputs are empirically most prevalent, Arabic, and Urdu are included to validate cross-family generalization beyond Indic scripts.

\subsection{E5: Orthographic Stress Test}

To test robustness under extreme script mismatch, we inject additional romanization into model hypotheses (10–50\% token replacement) while keeping references unchanged (Table~\ref{tab:e5-mixed}).

On FLEURS, average WER inflation from 0\%→50\% mixing was $\Delta$WER=0.234 across five Indic languages, whereas SN-WER increased only $\Delta$SN=0.158, attenuating 67\% of script-driven inflation. Attenuation ranged from 0.58 (Hindi) to 0.81 (Tamil).


These results confirm that SN-WER reduces artificial script inflation while preserving residual error sensitivity. CER remains appropriate for some languages and applications, especially where word segmentation is not standard; however, in this Indic romanization stress test, CER remains sensitive to script form (mean $\Delta$CER/$\Delta$WER$\approx$1.8) and does not isolate lexical recognition from script mismatch.

\begin{table}[h]
\centering
\begin{tabular}{l c c c}
\toprule
\textbf{Lang} & $\Delta$WER & $\Delta$SN & Attenuation \\
\midrule
bn & 0.1650 & 0.1002 & 0.607 \\
gu & 0.2362 & 0.1654 & 0.700 \\
hi & 0.2901 & 0.1693 & 0.584 \\
ta & 0.2437 & 0.1965 & 0.806 \\
or & 0.2151 & 0.1452 & 0.671 \\
\midrule
\textbf{Mean} & \textbf{0.2338} & \textbf{0.1578} & \textbf{0.674} \\
\bottomrule
\end{tabular}
\caption{Controlled 0\%→50\% hypothesis romanization. SN-WER attenuates 67.4\% of script-driven WER inflation.}
\label{tab:e5-mixed}

\end{table}

\subsection{E6: Lexical Sensitivity Control}
To verify lexical sensitivity, we inject controlled lexical substitutions (20--30\% token corruption) into hypotheses without altering script (Table~\ref{tab:e6-semantic}).

Across five Indic languages, WER increases by $\Delta \approx 0.08$--$0.14$, and SN-WER increases nearly identically (ratio $\approx 1.05$--$1.13$). Unlike script perturbations, SN-WER does not attenuate lexical errors, confirming that it does not simply lower scores.

Taken together, E5 and E6 validate SN-WER's intended behavior. Under artificial script perturbation (0--50\% hypothesis romanization), SN-WER attenuates 67\% of WER inflation across languages. In contrast, under lexical corruption (20--30\%), SN-WER tracks WER nearly identically (mean ratio $\approx 1.09$), showing that script normalization does not weaken sensitivity to true recognition errors.

\begin{table}[h]
\centering
\footnotesize
\setlength{\tabcolsep}{2pt}
\renewcommand{\arraystretch}{1.05}

\begin{tabular*}{\columnwidth}{@{\extracolsep{\fill}} l c c c @{}}
\toprule
\textbf{Lang} & $\boldsymbol{\Delta}$\textbf{WER} & $\boldsymbol{\Delta}$\textbf{SN} & \textbf{Ratio ($\boldsymbol{\Delta}$SN/$\boldsymbol{\Delta}$WER)} \\
\midrule
bn & 0.0785 & 0.0839 & 1.07 \\
gu & 0.1089 & 0.1230 & 1.13 \\
hi & 0.1391 & 0.1464 & 1.05 \\
ta & 0.1105 & 0.1236 & 1.12 \\
or & 0.1152 & 0.1262 & 1.10 \\
\midrule
\textbf{Mean} & \textbf{0.1093} & \textbf{0.1192} & \textbf{1.09} \\
\bottomrule
\end{tabular*}

\caption{Lexical corruption at 20--30\%. SN-WER tracks WER closely, indicating no attenuation of lexical errors.}
\label{tab:e6-semantic}
\end{table}

\subsection{E7: Statistical \& Adversarial Validation}

\begin{figure}[h]
\centering

    \centering
    \includegraphics[width=1\linewidth]{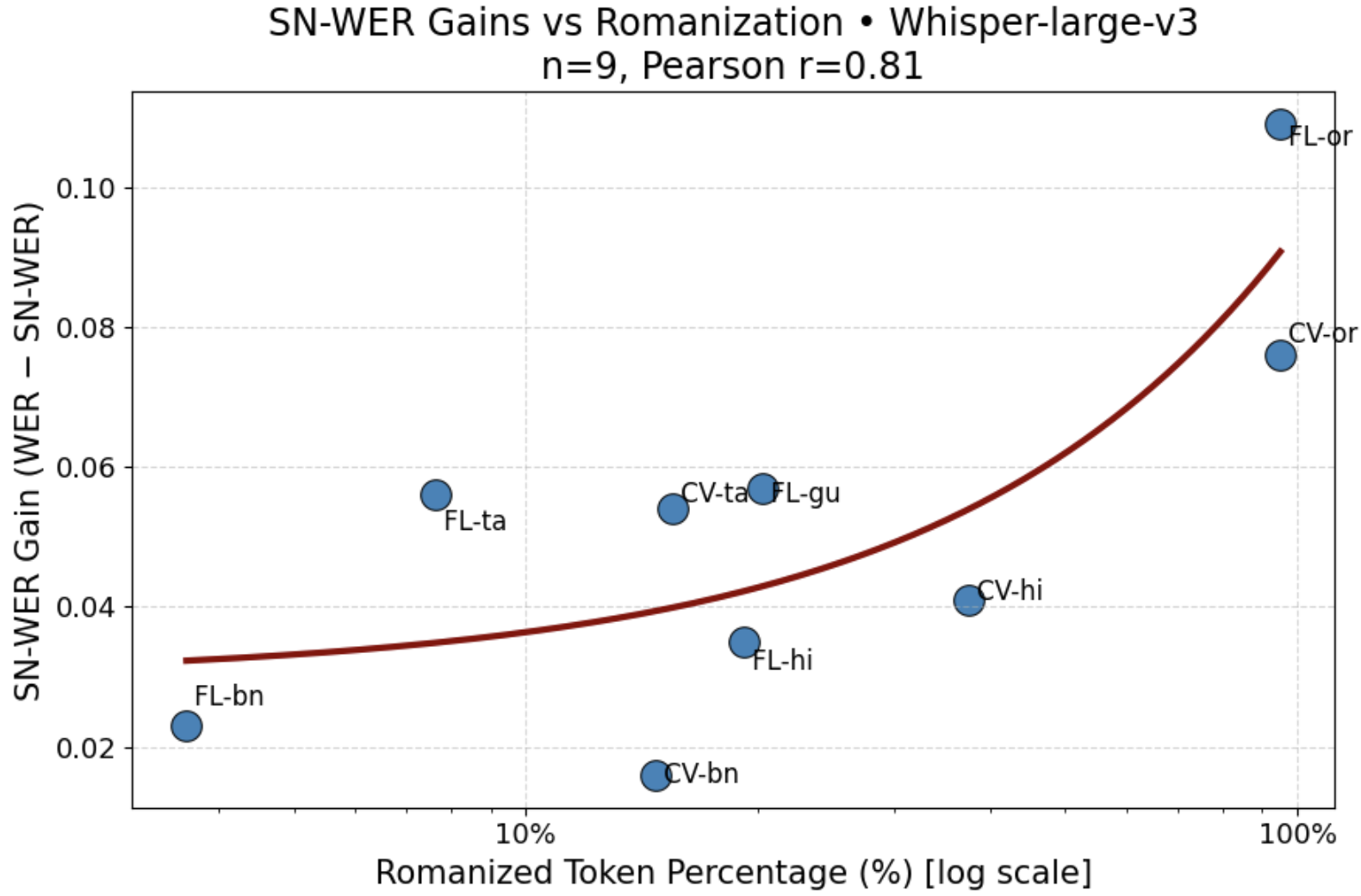}

    \label{fig:placeholder}

\caption{Correlation between romanization rate and $\Delta_{\text{WER}\to\text{SN-WER}}$}
\label{fig:romanization}
\vspace{-0.4cm}
\end{figure}

\textbf{Romanization correlation.}
Romanization rate is computed via Unicode block detection: tokens with majority Latin characters are marked romanized; others use language-specific script ranges. The magnitude of SN-WER correction correlates with baseline romanization rate ($r=0.81$ for Whisper-large), confirming that correction scales with script mismatch rather than arbitrary normalization effects (Fig.~\ref{fig:romanization}).

\textbf{Bootstrap confidence.}
Paired bootstrap resampling (1k samples) shows that WER--SN-WER score reductions are statistically significant ($p < 0.05$) across all five Indic languages, with CI widths $\leq 0.02$.

\textbf{Adversarial sanity.}
As a sanity check, we construct adversarial hypotheses by either randomly permuting token order or replacing content tokens with different same-script tokens. These perturbations should remain errors under any valid scoring method. SN-WER rises to $\approx 1.0$ (Table~\ref{tab:adv-sanity-full}), confirming that transliteration does not mask word-order or lexical errors.

These statistical and adversarial checks provide stronger evidence than isolated examples: the score reduction correlates with romanization rate, while adversarial lexical and word-order perturbations remain heavily penalized.

\begin{table}[h]
\centering
\footnotesize
\setlength{\tabcolsep}{2pt}
\renewcommand{\arraystretch}{1.05}

\begin{tabular*}{\columnwidth}{@{\extracolsep{\fill}} l c c c @{}}
\toprule
\textbf{Language} & \textbf{Base SN-WER} & \textbf{Shuffle} & \textbf{Substitute} \\
\midrule
Hindi    & 0.42 & 0.96 & 0.91 \\
Bengali  & 0.53 & 0.97 & 0.92 \\
Tamil    & 0.61 & 0.98 & 0.94 \\
Odia     & 0.49 & 0.97 & 0.93 \\
Gujarati & 0.46 & 0.96 & 0.92 \\
\bottomrule
\end{tabular*}

\caption{Adversarial sanity across languages: SN-WER penalizes lexical errors heavily.}
\label{tab:adv-sanity-full}
\end{table}

\subsection{Discussion}

Our experiments reveal three key patterns:

First, SN-WER reduces script-driven inflation on curated FLEURS: model gaps shrink by up to 12\%, indicating that part of the measured WER gap is due to script mismatch rather than lexical recognition failure.

Second, on noisy Common Voice, SN-WER often widens or preserves gaps, revealing true model fragility. Whisper-small's WER$>$1.0 from insertions is lowered, but real errors remain penalized.

Third, controlled perturbations clarify scope. Under artificial mixed-script stress (0--50\% romanization), WER inflation averages $\Delta$0.23 while SN-WER increases $\Delta$0.16, attenuating 67\% of script-driven inflation. Under lexical corruption (20--30\%), SN-WER tracks WER closely (ratio $\approx$1.09), confirming it does not weaken sensitivity to true recognition errors.

Evaluation checks further support reliability in the evaluated setting: transliterator invariance (mean disagreement $\approx$0.002), normalization stability ($\Delta{<}0.05$), strong romanization correlation ($r=0.81$), bootstrap significance, negligible collision rates ($<0.1\%$), and sanity tests (SN-WER$\to$1.0 under shuffles). Together, these show that SN-WER estimates script-mismatch effects while preserving lexical sensitivity.

SN-WER is most useful as a companion score for script-insensitive settings such as search, indexing, retrieval, intent classification, and downstream multilingual LLM processing. For user-facing transcripts, captions, subtitles, or educational applications, WER/CER should still be reported because correct script choice remains part of output quality.

\subsection{Conclusion}

We presented Script-Normalized WER (SN-WER), an evaluation-only companion score for estimating script-mismatch effects in multi-script ASR evaluation. SN-WER transliterates reference and hypothesis into a language-specific canonical script before computing WER, requiring no retraining, decoding changes, or additional labeled data. Across five Indic languages, two datasets, and three ASR models, SN-WER narrows inflated gaps on curated data while preserving sensitivity to genuine lexical errors on noisy and adversarial inputs. Controlled stress tests show 67\% attenuation of artificial romanization-induced WER inflation, while lexical corruption controls show near-identical sensitivity to true recognition errors. Additional Arabic and Urdu results suggest that the approach can extend beyond Indic when language-specific orthographic conventions are respected. Robustness checks show low transliterator disagreement, normalization stability, and negligible collision risk in the evaluated setting. SN-WER should be reported alongside WER/CER, not as a replacement, especially when the user-facing script is part of output quality. Future work will extend SN-WER to code-switching, language-specific scoring conventions, and downstream LLM-based speech applications.




\bibliography{mybib}

\end{document}